\documentclass[iicol]{sn-jnl}

\usepackage{graphicx}%
\usepackage{multirow}%
\usepackage{multicol}
\usepackage{amsmath,amssymb,amsfonts}%
\usepackage{amsthm}%
\usepackage{mathrsfs}%
\usepackage[title]{appendix}%
\usepackage{xcolor}%
\usepackage{textcomp}%
\usepackage{manyfoot}%
\usepackage{booktabs}%
\usepackage{algorithm}%
\usepackage{algorithmicx}%
\usepackage{algpseudocode}%
\usepackage{listings}%

\usepackage{array} 
\usepackage{subfigure}
\definecolor{reviewcolor}{RGB}{0, 150, 0}

\theoremstyle{thmstyleone}%
\theoremstyle{thmstyletwo}%

\theoremstyle{thmstylethree}%

\begin{document}

\title[FaceSaliencyAug: Mitigating Geographic, Gender and Stereotypical Biases via Saliency-Based Data Augmentation]{FaceSaliencyAug: Mitigating Geographic, Gender and Stereotypical Biases via Saliency-Based Data Augmentation}


\author*[1]{\fnm{Teerath} \sur{Kumar}}\email{teerath.menghwar2@mail.dcu.ie}

\author[2]{\fnm{Alessandra} \sur{Mileo}}\email{alessandra.mileo@dcu.ie}

\author[3]{\fnm{Malika} \sur{Bendechache}}\email{malika.bendechache@universityofgalway.ie}


\affil*[1]{\orgdiv{CRT-AI \& ADAPT Research Centre, School of Computing}, \orgname{Dublin City University},\orgaddress{ \city{Dublin 9}, \state{Dublin}, \country{Ireland}}}
 
\affil*[2]{\orgdiv{INSIGHT \& I-Form Research Centre, School of Computing}, \orgname{Dublin City University},\orgaddress{ \city{Dublin 9}, \state{Dublin}, \country{Ireland}}}

\affil[3]{\orgdiv{ADAPT \& Lero Research Centres,
School of Computer Science}, \orgname{University of Galway}, \orgaddress{\city{Galway}, \country{Ireland}}}


\abstract{Geographical, gender and stereotypical biases in computer vision models pose significant challenges to their performance and fairness. {In this study, we present an approach named FaceSaliencyAug aimed at addressing the gender bias in} {Convolutional Neural Networks (CNNs) and Vision Transformers (ViTs). Leveraging the salient regions} { of faces detected by saliency, the propose approach mitigates geographical and stereotypical biases } {in the datasets. FaceSaliencyAug} randomly selects masks from a predefined search space and applies them to the salient region of face images, subsequently restoring the original  image with masked salient region. {The proposed} augmentation strategy enhances data diversity, thereby improving model performance and debiasing effects. We quantify dataset diversity using Image Similarity Score (ISS) across five datasets, including Flickr Faces HQ (FFHQ), WIKI, IMDB, Labelled Faces in the Wild (LFW), UTK Faces, and Diverse Dataset. The proposed approach demonstrates superior diversity metrics, as evaluated by ISS-intra and ISS-inter algorithms. Furthermore, we evaluate the effectiveness of our approach in mitigating gender bias on CEO, Engineer, Nurse, and School Teacher datasets. We use the Image-Image Association Score (IIAS) to measure gender bias in these occupations. Our experiments reveal a reduction in gender bias for both CNNs and ViTs, indicating the efficacy of our method in promoting fairness and inclusivity in computer vision models.}




\maketitle

\section{Introduction}

\begin{figure*}
    \begin{center}
        \includegraphics[width=0.8\textwidth, height=0.23\textwidth]{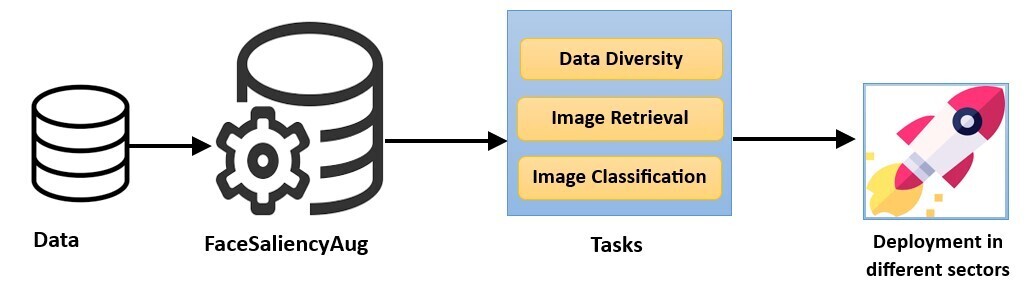}
    \end{center}
    \caption{Architectural diagram illustrating the complete methodology. Initially, FaceSaliencyAug data augmentation is applied to the dataset, followed by the utilization of the augmented data across three distinct tasks: data diversity, image retrieval, and image classification. Ultimately, the model can be deployed in various sectors.}
    \label{fig:architecture_diag}
\end{figure*}
\begin{figure*}
  
  \begin{center}
    \includegraphics[width=1.0\textwidth, height=0.52\textwidth]{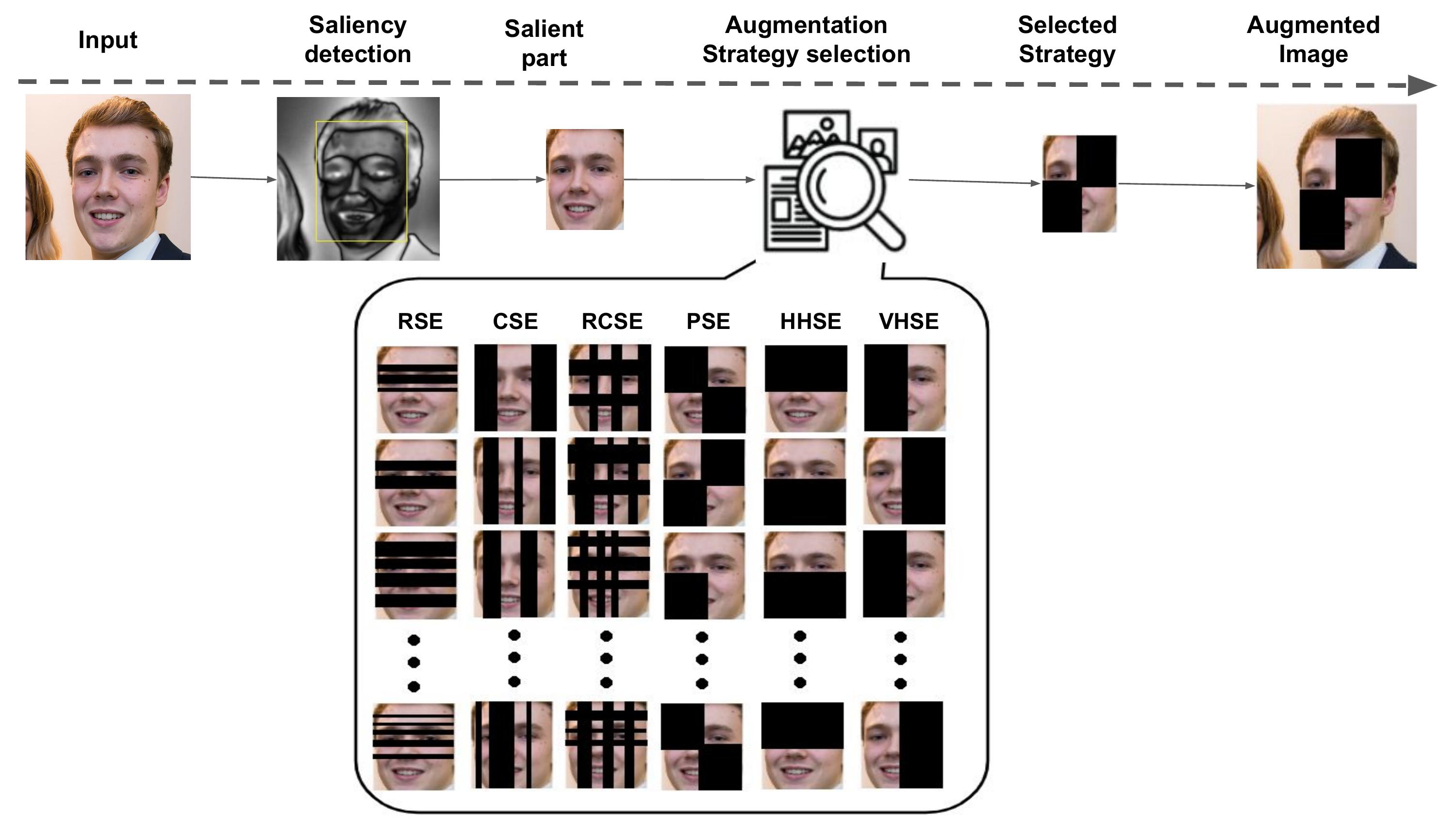}
\end{center}
  \caption{FaceSaliencyAug: Proposed approach to balance between complete object erasing and contextual information erasing, where  RSE, CSE, RCSE, PSE, HHSE and VHSE represent row slice erasing, column slice erasing, row-column saliency erasing, partial saliency erasing, horizontal half saliency erasing and vertical half saliency erasing, respectively.}

  \label{fig:methodology}
\end{figure*}

Computer vision models and their application have exhibited various social biases such as gender bias~\cite{buolamwini2018gender, birhane2021multimodal}, geographical bias~\cite{mandal2021dataset,mandal2023gender} and racial bias~\cite{buolamwini2018gender,karkkainen2021fairface}.  For example, the facial recognition system is less {accurate for people with dark skin} and also for women~\cite{buolamwini2018gender}. Another example is oxygen therapy; it is a common medical practice, monitored using a pulse oximeter, which measures oxygen levels in the blood by passing infrared light through the skin. However, this method tends to overestimate oxygen saturation in {non-white} patients, leading to under-detection of low oxygen levels, particularly among black patients, who are three times more likely to experience this discrepancy compared to white  patients~\cite{norori2021addressing}. Basic source of bias is  dataset used to train the model, which propagates the bias when deployed in real time.   For example, method for compiling datasets to train computer vision models is gathering images from the Internet, typically through search engines like Google or image hosting sites like Flickr (as seen in the case of FFHQ~\cite{karras2019ffhq}). However, this approach is prone to bias and can lead to the creation of biased datasets.  The auditing of social bias in visual datasets for faces has predominantly centered on two key parameters: race focusing on skin tone, and gender~\cite{mandal2021dataset,buolamwini2018gender,karkkainen2021fairface}, which is why facial region is important to debias the dataset.  To deal with these biases, several methods has been proposed~\cite{ zhang2020towards, kim2021biaswap, lee2021learning}. This work~\cite{zhang2020towards} investigates debasing in image classification task by generating adversarial examples to solve bias in training data distribution thus to increase model fairness. {Work by Kim et al.}~\cite{kim2021biaswap} introduces BiaSwap, which debiases deep neural networks without prior knowledge of bias types, using unsupervised sorting and style transfer to swap bias attributes between images. {Work by Lee et al.}~\cite{lee2021learning} introduced DiverseBias, a feature-level data augmentation technique for improving the debiasing of image classification models by synthesizing diverse bias-conflicting samples through disentangled representation learning.  In this paper, {w}e introduced a novel approach called FaceSaliencyAug to address {these biases}. {The proposed} method essentially identifies the salient region within a facial image, applies masking selected from a defined search space, and subsequently restores the original image with the masked salient region. {Moreover, 
the proposed architectural diagram outlines the methodology as shown in Fig.~\ref{fig:architecture_diag} it starts with Data, which undergoes FaceSaliencyAug. This augmented data is then utilized for various Tasks, including Diversity, Image Retrieval, and Model Training. Finally, the methodology is designed for Deployment in diverse applications such as Facial Recognition and Healthcare (more detail provided in section~\ref{real_world_application}.} The overview of our method is  shown in Fig.~\ref{fig:methodology}  Our work addresses below questions: 
\begin{itemize}
    \item To what extend does FaceSaliencyAug contribute to enhancing dataset diversity, as measured by the Image Similarity Score (ISS), across a range of datasets such as Flickr Faces HQ (FFHQ), WIKI, IMDB, Labelled Faces in the Wild (LFW), and UTK?
\item How does FaceSaliencyAug improve gender bias mitigation in CNNs and ViTs contributing to enhanced fairness in computer vision models across four occupation datasets?
\end{itemize}

Summary of our contribution as fellow: 
\begin{itemize}
    \item Introduction of FaceSaliencyAug, a pioneering approach targeting gender bias in CNNs and ViTs, while concurrently reducing geographical and stereotypical biases to enhance dataset diversity.
\item Development of a data augmentation method harnessing salient face regions to enrich dataset diversity, boost model performance, and effectively mitigate biases.
\item Validation of the proposed approach through the achievement of superior diversity metrics across {five} datasets, coupled with a substantial reduction in gender bias {across four occupation datasets}. This advancement underscores its role in advancing fairness and inclusivity in computer vision models.
\end{itemize}
\begin{figure*}[ht!]
\hfill
\subfigure[ Sample of Row Slice Erasing (RSE) visualisation]{ \includegraphics[width=0.48\textwidth, height=2.0cm]{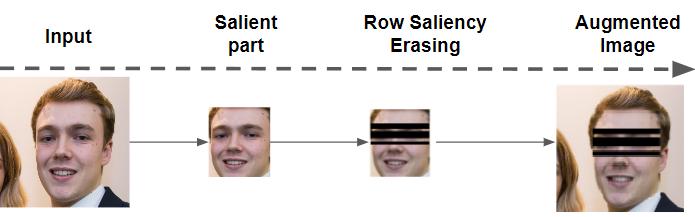}
    \label{fig:row_erasing}}
\hfill
\subfigure[Sample of  Column Slice Erasing (CSE) visualisation]{ \includegraphics[width=0.48\textwidth, height=2.0cm]{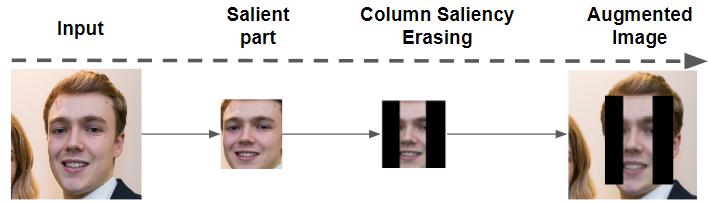}
    \label{fig:column_erasing}}

\medskip
\subfigure[Sample of  Row Column Slice Erasing (RCSE) visualisation]{  \includegraphics[width=0.48\textwidth, height=2.0cm]{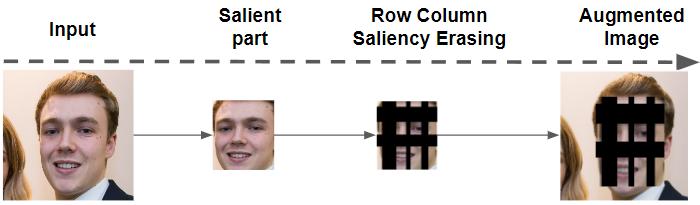}
    \label{fig:row_column}}
\hfill
\subfigure[Sample of  Partially Slice Erasing (PSE) visualisation]{ \includegraphics[width=0.48\textwidth, height=2.0cm]{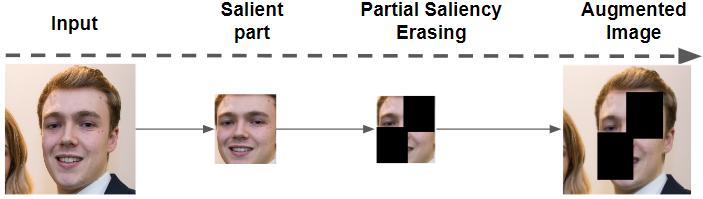}
    \label{fig:partial_erasing}}
\hfill

\medskip

\subfigure[Sample of  Horizontal Half Slice Erasing (HHSE) visualisation]{  \includegraphics[width=0.48\textwidth, height=2.0cm]{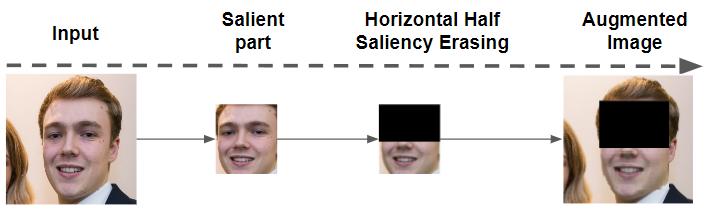}
    \label{fig:horizontal_half}}
\hfill
    \subfigure[Sample of  Vertical Half Slice Erasing (VHSE) visualisation]{   \includegraphics[width=0.48\textwidth, height=2.0cm]{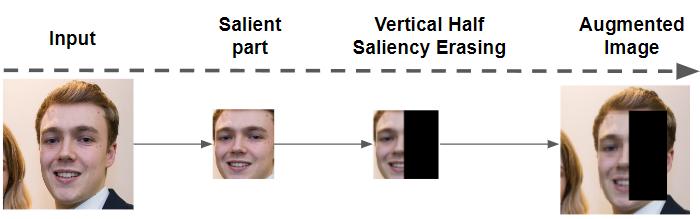}

    \label{fig:vertifical_half}}
\caption{The proposed augmentation strategies visualisation for the search space}
\label{fig:search_space}
\end{figure*}
\section{Related Work}
There have been several studies on bias in image domian~\cite{kumar2021binary, chandio2022precise, roy2023wildect, kumar2024image, kumar2024keeporiginalaugment, dixit2022trends, vavekanand2024cardiacnet, raj2024oxml, singh2024efficient,baea2021class}   
, audio domain~\cite{park2020search, kumar2020intra,khan2021robust, kumar2023audrandaug,turab2206investigating}, text domain~\cite{ kumarforged,khan2023sql,kumar2023advanced,kumar2023efficient,singh2301understanding,kumar2023understanding} and many more~\cite{aleem2022random,roy2022computer,raj2023understanding,singh2023deep, ranjbarzadeh2023me, khan2209sql},  and it is now attracting more attention than ever. Furthermore, as indicated by Mehrabi et al. \cite{mehrabi2021survey} in their review, historical, representational, {evaluation}, and sampling biases often occur- appearing as gender and racial biases equally. {The work} by Buolamwini and Gebru \cite{buolamwini2018gender} shows that, indeed, the current commercial facial recognition systems that detect and recognize faces of dark-skinned women tend to produce errors in their performance. Celis and Keswani \cite{celis2020implicit} have proposed a model {to check} socially biased image retrieval. They found that the search queries often carry the regional social, cultural and demographic characteristics thus generating photo representations that have biased characteristics. In order to deal with biases, researchers have made it possible to use curated datasets such as Fairface \cite{karkkainen2021fairface}, although there may be {left} biases in the datasets. A huge problem is represented by the numerous visual datasets that come from Western institutions, which are the First Cause of this bias \cite{mandal2021dataset}. The Internet-based {training, growing exponentially,} requires more extensive filtering and debiasing performance \cite{birhane2021multimodal}. We suggest a ground-breaking data augmentation method which will ensure dataset diversity to minimise the regional bias in image search thus enabling one to remove the race, cultural, or stereotypical biases in a facial dataset. Besides, we quantify the worth of our approach in different versions of both CNN and ViT architectures, applying the MSA mechanism in ViT to catch the long distant dependencies and widening the receptive field.

Several image data augmentation methods have been proposed to address biases \cite{lim2023biasadv,kumar2023image,zhang2020towards,kim2021biaswap,lee2021learning}, including saliency-based approaches \cite{uddin2020saliencymix, choi2021salfmix}. Zhang et al. \cite{zhang2020towards} generate adversarial examples for bias reduction in image classification, enhancing model fairness. {Biaswap ~\cite{kim2021biaswap} debiases neural networks use unsupervised sorting and style transfer, effectively altering the biases present in training data. While this method has shown promise, it lacks exploration of how partial erasure of salient regions can further enhance debiasing. Lee et al.~\cite{lee2021learning} have proposed a technique that improves debiasing through the synthesis of diverse bias-conflicting samples. This approach is valuable as it introduces variability in training data, but it may not adequately address the underlying biases in the salient features of images. Similarly, Uddin et al.~\cite{uddin2020saliencymix} utilize saliency detection in source images, transferring salient features to non-salient areas in target images. Although this method successfully manipulates image content, it does not consider the effects of partially erasing salient regions with various shapes, which could potentially lead to better bias mitigation outcomes. Choi et al.~\cite{choi2021salfmix} introduce single-image saliency data augmentation, allowing for the transfer of salient regions to non-salient areas. However, like the previous methods, it falls short of exploring the impact of partial erasure on biases, particularly in facial regions.
Moreover, a critical examination of these methods reveals a gap in addressing geographical and stereotypical biases. Understanding these biases is essential in the context of localized and personalized search results. Research indicates that implicit AI biases contribute to geographical bias, where the localization and personalization of search engines can perpetuate stereotypes, such as the prevalent stereotype of the "white, middle-aged female nurse" in Western societies~\cite{mandal2021dataset}.}

\section{Proposed Approach}\label{proposed_method}
In this section, we introduce six data augmentation strategies for the search space and outline the proposed approach based on this search space.

\subsection{Search Space - Data Augmentation}\label{search_space}
The search space encompasses six proposed data augmentation techniques, each detailed below. It's worth noting that these augmentation methods operate on the salient region of the image, which is detected using methodologies outlined in previous works.

\subsubsection{Row Slice Erasing (RSE)}\label{row_slice}
{RSE} technique augments the salient region $\mathit{x}$ of image $I$ by element-wise multiplication with a binary mask $M$. The mask $M$ contains values of 0 or 1, denoting exclusion or inclusion of pixels, respectively. Slices of size $S$ are randomly selected from the salient region to generate the binary mask. Horizontal slices of the mask are filled alternately with 0's and 1's. Refer to {Fig.~\ref{fig:row_erasing}} for visualization.
\begin{equation}
 \tilde{x} =  x \odot M
 \label{eq:saliencyRow}   
\end{equation}

\subsubsection{Column Slice Erasing (CSE)}\label{cse}
Similar to RSE, CSE involves applying augmentation to the salient region $\mathit{x}$ of the image $I$. The augmented salient part $\tilde{x}$ is obtained using a binary mask $M$ generated by selecting slices of size $S$ from the salient region. However, in this strategy, vertical slices of the mask are alternately filled with 0's and 1's. See {Fig.~\ref{fig:column_erasing}} for a visualization.

\subsubsection{Row Column Slice Erasing (RCSE)}
RCSE combines RSE~\ref{row_slice} and CSE~\ref{cse} techniques. RSE and CSE are sequentially applied. RCSE is illustrated in {Fig.~\ref{fig:row_column}}.
\subsubsection{Partially Saliency Erasing (PSE)}
PSE divides the salient region into four parts and randomly erases one or more squares ({Fig.~\ref{fig:partial_erasing}}). Mathematically, a binary mask $M$ is split into four equal parts, with a random number of parts filled randomly with 0's or 1's. Element-wise multiplication generates the augmented image $\tilde{x}$ (Equation~\ref{eq:saliencyRow}).
\subsubsection{Horizontal Half Saliency Erasing (HHSE)}
HHSE horizontally divides the salient region into two parts and randomly erases one part ({Fig.~\ref{fig:horizontal_half}}). The mask $M$ is partitioned into two segments, with one filled with 0's and the other with 1's. This process generates the augmented image $\tilde{x}$ through element-wise multiplication (Equation~\ref{eq:saliencyRow}).
\subsubsection{Vertical Half Saliency Erasing (VHSE)}
VHSE vertically divides the salient region into two parts and randomly erases one part ({Fig.~\ref{fig:vertifical_half}}). The mask $M$ is split vertically into two equal-sized segments, with one filled with 0's and the other with 1's. This process generates the augmented image $\tilde{x}$ via element-wise multiplication (Equation~\ref{eq:saliencyRow}).

\subsection{FaceRandAug:}

Given an input image $I$ and a salient region $\mathit{x}$ detected within $I$, FaceRandAug randomly selects one of the following erasing strategies from the given data augmentation list with equal probability:
RSE, CSE, RCSE, PSE, HHSE and VHSE. \\  
Let $S$ denote the selected erasing strategy. The augmented salient region $\tilde{x}$ is obtained as follows:

\[
\tilde{x} = 
\begin{cases}
\text{Erased Salient Region using } S, & \text{if } S \text{ is selected} \\
\mathit{x}, & \text{otherwise}
\end{cases}
\]

where $\tilde{x}$ represents the augmented salient region. The selection process of FaceRandAug ensures that each erasing strategy has an equal chance of being selected, similar to the searching process of RandAug~\cite{cubuk2020randaugment}. This method provides a flexible way to incorporate various erasing strategies into the data augmentation pipeline for facial mask selection.

\section{Experiments}
In this section, we discuss experiments settings, results and insights. 


\begin{table*}[hpt!]
\centering

\caption{Image Similarity score across all possible queries. Baseline results are taken from~\cite{mandal2021dataset}.}
\label{tab:intra_cross_iss}
\begin{tabular}{clcccc}
\toprule 
\multirow{2}{*}{\textbf{Query}} & \multirow{2}{*}{\textbf{Language Location Pair}} & \multicolumn{2}{c}{\textbf{ISS$_{Intra}$}} & \multicolumn{2}{c}{\textbf{ISS$_{cross}$}} \\

 & & \textbf{Baseline} & \textbf{Ours} & \textbf{Baseline} & \textbf{Ours} \\
\midrule 
\multirow{9}{*}{CEO} & Arabic-West Asia \& North Africa & 0.899012 & 0.906985 & \multirow{9}{*}{0.984683} & \multirow{9}{*}{0.987136} \\

 & English-North America & 0.968974 & 0.976045 & & \\

 & English-West Europe & 0.929469 & 0.957394 & & \\

 & Hindi-South Asia & 0.997845 & 0.994985 & & \\

 & Indonesian-SE Asia & 0.983675 & 0.985087 & & \\

 & Mandarin-East Asia & 0.989452 & 0.994088 & & \\

 & Russian-East Europe & 0.959661 & 0.964493 & & \\

 & Spanish-Latin America & 0.974743 & 0.974597 & & \\

 & Swahili-Sub Saharan Africa & 0.977119 & 0.975551 & & \\
\midrule 
\multirow{9}{*}{Engineer} & Arabic-West Asia \& North Africa & 0.98639 & 0.982757 & \multirow{9}{*}{0.993904} & \multirow{9}{*}{0.994201} \\

 & English-North America & 0.988344 & 0.991277 & & \\

 & English-West Europe & 1.000911 & 1.001757 & & \\

 & Hindi-South Asia & 1.003149 & 0.994307 & & \\

 & Indonesian-SE Asia & 0.987191 & 0.990045 & & \\

 & Mandarin-East Asia & 0.991146 & 0.987163 & & \\

 & Russian-East Europe & 1.007155 & 1.000498 & & \\

 & Spanish-Latin America & 0.984955 & 0.986797 & & \\

 & Swahili-Sub Saharan Africa & 0.983727 & 0.986656 & & \\
\midrule
\multirow{9}{*}{Nurse} & Arabic-West Asia \& North Africa & 1.002607 & 0.995654 & \multirow{9}{*}{0.989952} & \multirow{9}{*}{0.990378} \\

 & English-North America & 0.971564 & 0.973874 & & \\

 & English-West Europe & 0.99561 & 0.992313 & & \\

 & Hindi-South Asia & 0.984535 & 0.987963 & & \\

 & Indonesian-SE Asia & 0.975914 & 0.979673 & & \\

 & Mandarin-East Asia & 0.98904 & 0.993333 & & \\

 & Russian-East Europe & 0.997979 & 0.996215 & & \\

 & Spanish-Latin America & 1.000587 & 0.993549 & & \\

 & Swahili-Sub Saharan Africa & 0.958532 & 0.972825 & & \\
\midrule

\multirow{9}{*}{Politician} & Arabic-West Asia \& North Africa & 0.977348 & 0.979589 & \multirow{9}{*}{0.983637} & \multirow{9}{*}{0.987169} \\

 & English-North America & 0.995927 & 0.999067 & & \\

 & English-West Europe & 0.979358 & 0.981664 & & \\

 & Hindi-South Asia & 0.979915 & 0.980589 & & \\

 & Indonesian-SE Asia & 0.972307 & 0.992733 & & \\

 & Mandarin-East Asia & 0.976251 & 0.973902 & & \\

 & Russian-East Europe & 0.93835 & 0.965788 & & \\

 & Spanish-Latin America & 0.988452 & 0.989963 & & \\

 & Swahili-Sub Saharan Africa & 0.943626 & 0.952849 & & \\
\midrule
\multirow{9}{*}{\begin{tabular}{l} School \\ Teacher \end{tabular}} & Arabic-West Asia \& North Africa & 1.014298 & 1.013607 & \multirow{9}{*}{0.990403} & \multirow{9}{*}{0.989746} \\

 & English-North America & 0.997715 & 0.988912 & & \\

 & English-West Europe & 0.940142 & 0.956337 & & \\

 & Hindi-South Asia & 1.000047 & 0.995507 & & \\

 & Indonesian-SE Asia & 0.985991 & 0.990466 & & \\

 & Mandarin-East Asia & 1.00862 & 1.004755 & & \\

 & Russian-East Europe & 0.976169 & 0.962497 & & \\

 & Spanish-Latin America & 0.965902 & 0.961019 & & \\

 & Swahili-Sub Saharan Africa & 0.985919 & 0.990466 & & \\
\bottomrule
\end{tabular}
\end{table*}

\begin{table*}[hpt!]
\centering
\caption{ISS$_{intra}$ of Datasets and baseline result originate from~\cite{mandal2021dataset}.}
\label{tab:iss_intra}
\begin{tabular}{lccc}
\toprule
Dataset & Baseline & RSMDA~\cite{kumar2023rsmda} & Ours \\
\midrule
FFHQ~\cite{karras2019ffhq} & $0.9940$ & $0.9935$ & \textbf{0.9939} \\
Diverse Dataset~\cite{mandal2021dataset} & $0.9895$ & $0.9900$ & \textbf{0.9905} \\
WIKI~\cite{rothe2015dex} & $0.9786$ & $0.9800$ & \textbf{0.9832} \\
IMDB~\cite{rothe2015dex} & $0.9661$ & $0.9700$ & \textbf{0.9717} \\
LFW~\cite{learned2016labeled} & $0.9536$ & \textbf{0.9550} & 0.9539 \\
UTK~\cite{zhang2017age} & $0.9418$ & $0.9440$ & \textbf{0.9481} \\
\bottomrule
\end{tabular}
\end{table*}

\begin{table*}[hpt!]
    \centering
        \caption{Image Similarity score across all possible queries. Baseline results are taken from~\cite{mandal2021dataset}.}
    \label{tab:iss_scores}
\begin{tabular}{lcccc}
\toprule & \multicolumn{2}{c}{ ISS$_{intra}$ } & \multicolumn{2}{c}{ ISS$_{cross}$} \\
\midrule \multicolumn{1}{c}{ Query } & Baseline & Ours & Baseline & Ours \\
\midrule
 CEO & 0.9644 & \textbf{0.9699} & 0.9846 & \textbf{0.9871} \\
 Engineer & \textbf{0.9925} & 0.9913 & 0.9939 & \textbf{0.9942} \\
Nurse & 0.9862 & \textbf{0.9873} & 0.990 & \textbf{0.9904} \\
Politician & 0.9724 & \textbf{0.9796} & 0.9836 & 0.9872 \\
School Teacher & \textbf{0.9860} & 0.9848 & \textbf{0.9904} & 0.9897 \\
Mean Value & 0.9803 & \textbf{0.9826} & 0.9885 & \textbf{0.9897} \\
\bottomrule
\end{tabular}
\vspace{-1pt} 

\end{table*}

\begin{table*}[hpt!]
\centering
\caption{Average Image-Image Association Scores (IIAS) for CNNs and ViTs. Positive values indicate bias towards men, negative towards women. Total absolute IIAS reflects bias magnitude. Our approach reduces gender bias, highlighted in red. {Baseline result and dataset were taken from \cite{mandal2023biased, mandal2023gender}}}
\label{tab:gender_bias}
\begin{tabular}{lcccccccc}
\toprule  \multicolumn{9}{c}{\textbf{Masked}} \\
\midrule  Class  & \multicolumn{4}{c}{ Biased } & \multicolumn{4}{c}{ Unbiased } \\
\midrule
 & $\underset{\text { Baseline }}{\mathrm{CNN}}$ & 
$\underset{\text { Ours }}{\mathrm{CNN}}$ & 
$\underset{\text { Baseline }}{\mathrm{ViT}}$
 &  
$\underset{\text { Ours }}{\mathrm{ViT}}$
 & $\underset{\text { Baseline }}{\mathrm{CNN}}$ & 
$\underset{\text { Ours }}{\mathrm{CNN}}$ & 
$\underset{\text { Baseline }}{\mathrm{ViT}}$
 &  
$\underset{\text { Ours }}{\mathrm{ViT}}$
 
 \\
\midrule CEO & 0.059 & 0.005 & 0.1 & 0.010 & 0.26 & 0.007 & 0.02 & 0.007 \\
\hline Engineer & 0.23 & 0.018 & 0.14 & 0.019 & 0.36 & 0.005 & 0.17 & 0.021 \\
 Nurse & -0.14 & -0.021 & -0.35 & -0.0214 & -0.05 & -0.007 & -0.2 & -0.018 \\
 School Teacher & -0.17 & -0.004 & -0.15 & -0.027 & -0.12 & -0.004 & -0.05 & -0.013 \\
 \multirow[t]{2}{*}{ Total IIAS (abs) } & 0.599 & 0.048 & 0.74 & 0.077 & 0.79 & 0.023 & 0.44 & 0.059 \\ 
  \midrule
 Bias reduction & \multicolumn{2}{c}{\textcolor{red}{$\sim 12$ times $\downarrow$ }} & \multicolumn{2}{c}{\textcolor{red}{$\sim 10$ times $\downarrow$ }} & \multicolumn{2}{c}{\textcolor{red}{$\sim 34$ times$\downarrow$ }} & \multicolumn{2}{c}{\textcolor{red}{$\sim 8$ times$\downarrow$ }} \\
\toprule
  \multicolumn{9}{c}{ \textbf{Unmasked}} \\
\midrule
  Class  & \multicolumn{4}{c}{ Biased } & \multicolumn{4}{c}{ Unbiased } \\
\midrule
 &  $\underset{\text { Baseline }}{\mathrm{CNN}}$ & 
$\underset{\text { Ours }}{\mathrm{CNN}}$ & 
$\underset{\text { Baseline }}{\mathrm{ViT}}$
 &  
$\underset{\text { Ours }}{\mathrm{ViT}}$
 & $\underset{\text { Baseline }}{\mathrm{CNN}}$ & 
$\underset{\text { Ours }}{\mathrm{CNN}}$ & 
$\underset{\text { Baseline }}{\mathrm{ViT}}$
 &  
$\underset{\text { Ours }}{\mathrm{ViT}}$
 \\
 \midrule
 CEO & 0.050 & 0.022 & 0.17 & 0.004 & 0.07 & 0.003 & 0.06 & 0.002 \\
 Engineer & 0.180 & 0.028 & 0.19 & 0.008 & 0.04 & 0.016 & 0.21 & 0.003 \\
 Nurse & -0.21 & -0.054 & -0.21 & -0.008 & -0.06 & -0.002 & -0.17 & -0.005 \\
 School Teacher & -0.02 & -0.025 & -0.4 & -0.006 & -0.04 & -0.0002 & -0.14 & -0.001 \\

 \multirow[t]{2}{*}{ Total IIAS (abs) } & 0.46 & 0.129 & 0.97 & 0.026 & 0.21 & 0.0212 & 0.58 & 0.011 \\
  \midrule
 Bias reduction & \multicolumn{2}{c}{\textcolor{red}{$\sim 4$ times $\downarrow$}} & \multicolumn{2}{c}{\textcolor{red}{$\sim 37$ times$\downarrow$ }} & \multicolumn{2}{c}{\textcolor{red}{$\sim 10$ times$\downarrow$ }} & \multicolumn{2}{r}{ \textcolor{red}{$\sim 53$ times$\downarrow$ }} \\
\bottomrule
\end{tabular}

\end{table*}

\subsection{Experimental Setting}
For measuring data diversity in terms of geographical and stereotypical biases, we use two variants of {the} Image Similarity Score (ISS) namely ISS$_{Intra}$ and ISS$_{Cross}$~\cite{mandal2021dataset}. ISS$_{Intra}$ measures data diversity within dataset and ISS$_{Cross}$ measures data diversity among different {datasets}~\cite{mandal2021dataset}. Five datasets (i.e: Flickr Faces HQ (FFHQ), WIKI, IMDB, Labelled Faces in the Wild (LFW), UTK Faces and Diverse Dataset) with same setting as used in work~\cite{mandal2021dataset}.

For measuring gender bias in CNNs and ViTs, we use Image-Image Association Score (IIAS)~\cite{mandal2023multimodal,mandal2023biased}. IIAS shows bias by comparing how similar the attributes representing gender in images using cosine similarity are to the images depicting specific concepts~\cite{mandal2023multimodal}. We employe the same setting as used in ~\cite{mandal2023gender}, four CNN models; (VGG16, ResNet152, Inceptionv3, and Xception), four ViT models (ViT B/16, B/32, L/16, and L/32). Models training {settings are} same as in the work ~\cite{mandal2023gender}. For CNNs, layers were frozen, then custom layers were trained for 50 epochs. Later, some layers were unfrozen and trained for 50 more epochs. For ViTs, layers were first frozen and trained for 100 epochs, then all layers were trained for 50 epochs with a low learning rate. 
A gender bias study collected images from Google searches using job terms. They made two training sets: equal gender and biased. Test set stayed balanced. Training: 7,200 images (900/category). Test: 1,200 images (300/category, 150/gender)~\cite{mandal2023gender,  mandal2023biased}. {We also} used separate men and women datasets for evaluation~\cite{mandal2023gender}. We integrated our approach by adding our data augmentation approach in transformations.  {The reason of choosing different profession datasets is that the categories encompass both masked and unmasked samples, providing an opportunity to evaluate the generalization capability of our proposed approach effectively. Lastly, the stereotypical biases associated with these roles, such as the perception of CEOs as predominantly male and nurses as predominantly female, underscore the significance of our study in addressing and mitigating biases in computer vision model}.

\begin{figure*}[hpt!]
    \centering
    \includegraphics[width=1.0\textwidth, height=10cm, trim= 2cm 0cm 0cm 0cm]{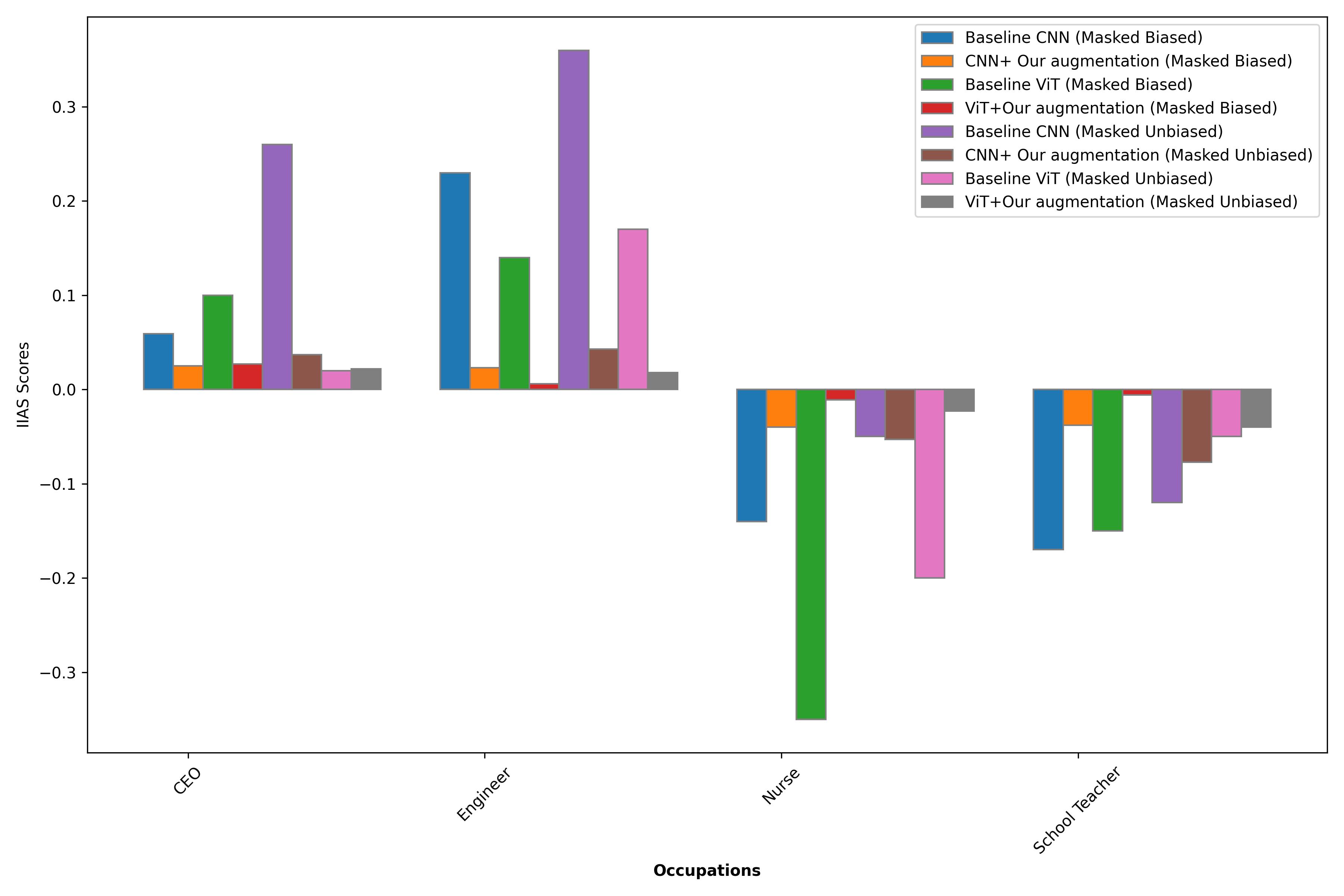}
    \caption{Comparison of our approach for gender bias reduction in CNN and ViT- Masked Scienario. }
    \label{fig:masked_biased}
    \vspace{2.5cm}
    \centering
    \includegraphics[width=1.0\textwidth, height=10cm, trim= 2cm 0cm 0cm 0cm]{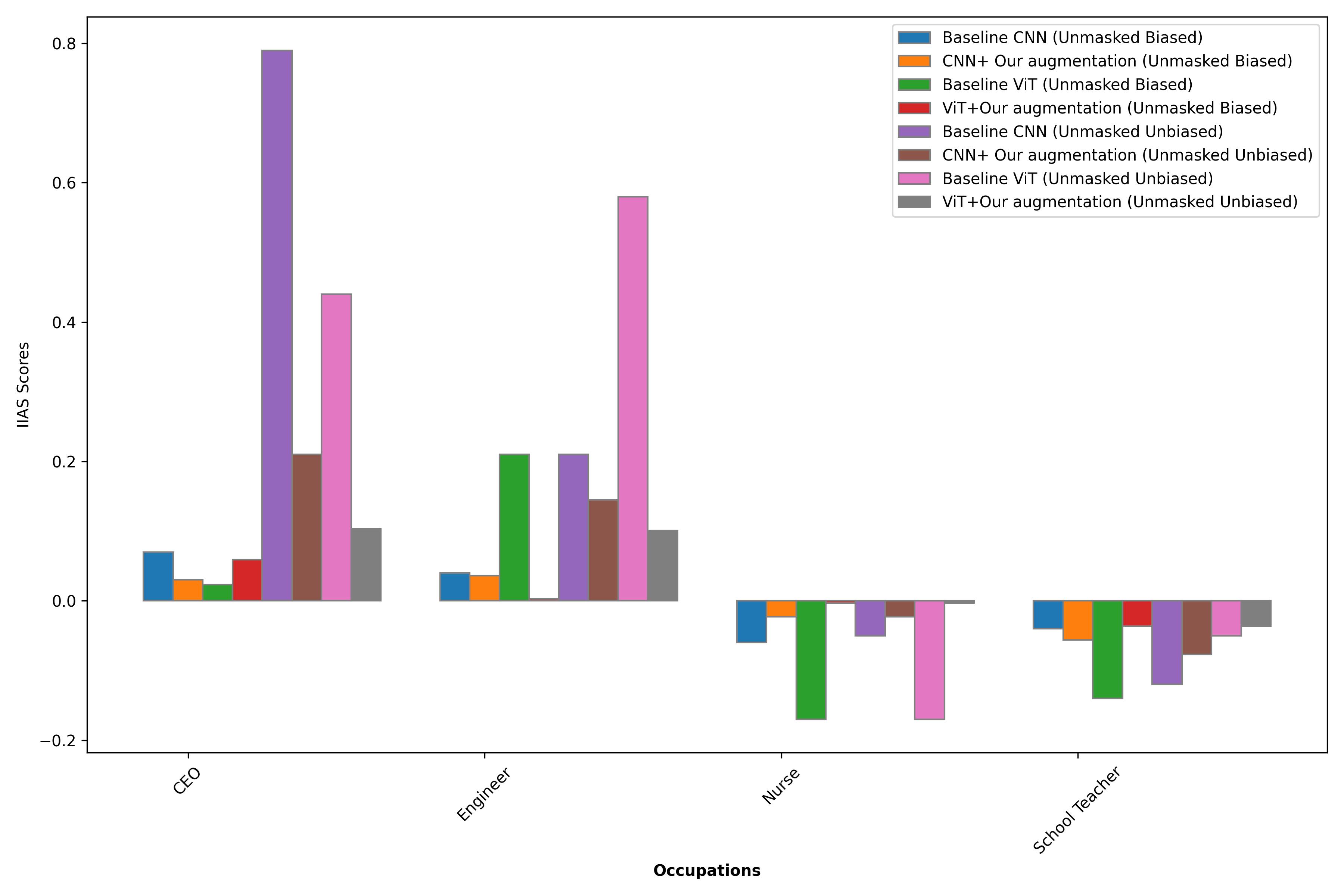}
    \caption{Comparison of our approach for gender bias reduction in CNN and ViT- Unmasked Scienario. }
    \label{fig:unmasked_biased}
\end{figure*}

\subsection{Result}

We {performed} experiments on two tasks: measuring dataset diversity across five datasets and evaluating gender diversity in four occupation datasets. For dataset diversity, we calculated Intra-dataset Image Similarity Score (ISS$_{intra}$), demonstrating enhanced diversity across all datasets (Table \ref{tab:iss_intra}), {and the proposed also shown superior performance than RSMDA~\cite{kumar2023rsmda} except for LFW dataset.} For a more detailed analysis of query combinations with language-location pairs, Table~\ref{tab:intra_cross_iss}
 presents the ISS$_{intra}$ for various queries across different language-location pairs. We compare the baseline ISS (intra and across) values with our proposed approach. The queries include CEO, Engineer, Nurse, Politician, and School Teacher, each evaluated with multiple language-location pairs. Overall, our proposed approach demonstrates higher diversity, as measured by ISS scores, as shown in Table~\ref{tab:iss_scores}.  Additionally, we also analyzed Inter-dataset Image Similarity Score (ISS$_{inter}$) and ISS$_{intra}$ for five occupation datasets. Our approach generally showed greater diversity, except for the school teacher dataset as shown in {Table~\ref{tab:iss_scores}}, potentially due to a female bias identified by Mandal et al. (2023) \cite{mandal2023gender}. Particularly, our approach consistently outperformed the baseline for "CEO" and "Engineer" occupations in both ISS$_{intra}$ and ISS$_{cross}$. Additionally, across various datasets, our approach achieved slightly better results compared to the baseline. Moreover, our approach yielded higher mean ISS scores across all queries, indicating its effectiveness in enhancing diversity. Overall, our approach achieved impressive diversity scores. 



In addition, we assessed gender bias using the Image-Image Association Scores (IIAS) on both masked (obscured) and unmasked (non-obscured) data, utilizing different CNNs and ViTs. Our proposed approach consistently reduced bias across all scenarios, with particularly significant reductions observed in the unmasked dataset. Notably, when the data was well-balanced (Unbiased) in terms of gender, our approach achieved an approximately 53-fold reduction in bias. Furthermore, our approach demonstrated superior performance across all classes compared to the baseline, as indicated by the larger reductions in bias for the "Ours" columns compared to the "Baseline" columns. Additionally, while our approach exhibited substantial reductions in bias across all occupations compared to the baseline, the magnitude of reduction varied across occupations and data types. For instance, the reduction in bias for the "Nurse" occupation was more pronounced in the unmasked dataset compared to the "School Teacher" occupation. Furthermore, our approach achieved particularly pronounced bias reduction for ViTs compared to CNNs, highlighting its effectiveness across different model architectures. While the reduction in bias was slightly less pronounced in the masked dataset compared to the unmasked dataset, our approach still exhibited substantial improvements. Overall, our approach proved highly effective in reducing gender bias, as highlighted in red in Table \ref{tab:gender_bias}.

{\subsubsection{Masked Bias reduction} In Fig.~\ref{fig:masked_biased}, the results reveal a clear difference in bias reduction between models for the masked scenario, where gender-specific features are hidden. For occupations such as CEO and Engineer, the Baseline CNN (Masked Biased) and Baseline ViT (Masked Biased) models display positive IIAS scores, indicating the presence of gender bias. The Baseline ViT (Masked Biased) model, in particular, exhibits higher bias in these occupations, with significantly positive scores compared to the CNN models. When applying our augmentation method, there is a noticeable reduction in bias. The CNN + Our Augmentation (Masked Biased) and ViT + Our Augmentation (Masked Biased) models show lower IIAS scores, indicating that the proposed augmentation strategy effectively reduces the bias in these occupations. Specifically, the ViT + Our Augmentation (Masked Biased) model significantly outperforms the Baseline ViT, reducing bias in both CEO and Engineer occupations. For the Nurse and School Teacher occupations, which typically reflect higher stereotypical associations, the bias is much more pronounced in the Baseline ViT (Masked Biased) model. However, with our augmentation, particularly in the ViT + Our Augmentation (Masked Biased) model, the bias is substantially reduced, bringing the IIAS scores closer to zero. This demonstrates that our augmentation method is effective in masking gendered features and reducing stereotypical associations in these biased roles. The CNN + Our Augmentation (Masked Biased) model similarly reduces bias, though the effect is more moderate compared to the ViT-based model.

\subsubsection{Unmasked Bias reduction}

In Fig.~\ref{fig:unmasked_biased}, where the unmasked scenario allows for gender-specific visual cues, the IIAS scores across all occupations are generally higher, reflecting stronger gender biases when the models have access to these cues. For CEO and Engineer, the Baseline CNN (Unmasked Biased) and Baseline ViT (Unmasked Biased) models show significant positive scores, indicating that bias is more evident when gender-related features are not masked. Once again, our augmentation method shows substantial improvements. The CNN + Our Augmentation (Unmasked Biased) and ViT + Our Augmentation (Unmasked Biased) models reduce the biases for CEO and Engineer, with ViT + Our Augmentation (Unmasked Biased) being particularly effective in lowering the IIAS scores, though the scores are higher than in the masked scenario. This indicates that while the augmentation reduces bias, the effect is slightly diminished in the unmasked scenario compared to the masked one. For the Nurse and School Teacher occupations, the Baseline ViT (Unmasked Biased) model shows the highest bias, with very negative IIAS scores, reflecting strong stereotypical associations. However, the ViT + Our Augmentation (Unmasked Biased) model effectively reduces these biases, significantly lowering the IIAS scores. The CNN + Our Augmentation (Unmasked Biased) model also reduces bias but does not perform as well as the ViT-based model. Overall, these results suggest that the augmentation method is effective in reducing bias in the unmasked scenario, though masking gender-specific features (as seen in Fig.~\ref{fig:masked_biased}) yields stronger bias mitigation.}

\begin{table*}[ht]
\footnotesize 
\centering
\caption{Accuracy performance comparison of the proposed approaches with the existing and relevant approaches on CIFAR10 and CIFAR100 datasets. Highlighted blue is the best performance }
\label{CIFAR_accuracies}
\begin{tabular}{ccccccc}
\toprule
\multicolumn{1}{l}{\textbf{Methods}}       & \multicolumn{1}{l}{\textbf{{ResNet32}}}              & \multicolumn{1}{l}{\textbf{ResNet44}}   & \multicolumn{1}{l|}{\textbf{ResNet56}}   & \multicolumn{1}{l}{\textbf{{ResNet32}}}              & \multicolumn{1}{l}{\textbf{ResNet44}}   & \multicolumn{1}{l}{\textbf{ResNet56}}  \\ \midrule
& \multicolumn{3}{c|}{\textbf{CIFAR10}}    &     \multicolumn{3}{c}{\textbf{CIFAR100}}                \\ \midrule

\multicolumn{1}{l}{Basline}  & \multicolumn{1}{c}{93.59 $\pm$ 0.06}  & \multicolumn{1}{c}{94.47 $\pm$ 0.08 }  & \multicolumn{1}{c|}{94.69 $\pm$ 0.07}    & \multicolumn{1}{c|}{71.50 $\pm$ 0.37}  & \multicolumn{1}{c}{74.73 $\pm$ 0.21  }  & \multicolumn{1}{c}{ 73.71 $\pm$ 0.28}               \\ 

\multicolumn{1}{l}{Dropout~\cite{srivastava2014dropout,liu2022focuseddropout}}   & \multicolumn{1}{c}{-}  & \multicolumn{1}{c}{-}  & \multicolumn{1}{c|}{93.74 $\pm$ 0.11}  & \multicolumn{1}{c}{-}  & \multicolumn{1}{c}{-}  & \multicolumn{1}{c}{73.81  $\pm$ 0.27 }                       \\ 

\multicolumn{1}{l}{S-Dropout~\cite{liu2022focuseddropout}}  & \multicolumn{1}{c}{-}  & \multicolumn{1}{c}{-}  & \multicolumn{1}{c|}{94.28 $\pm$ 0.20}  &  \multicolumn{1}{c}{-}  & \multicolumn{1}{c}{-}  & \multicolumn{1}{c}{74.97 $\pm$ 0.16 }                       \\ 

\multicolumn{1}{l}{DropBlock~\cite{ghiasi2018dropblock,liu2022focuseddropout}}   & \multicolumn{1}{c}{-}  & \multicolumn{1}{c}{-}  & \multicolumn{1}{c|}{94.01 $\pm$ 0.08}  & \multicolumn{1}{c}{-}  & \multicolumn{1}{c}{-}  & \multicolumn{1}{c}{73.92 $\pm$ 0.10  }                       \\

\multicolumn{1}{l}{F-Dropout~\cite{liu2022focuseddropout}}   & \multicolumn{1}{c}{-}  & \multicolumn{1}{c}{-}  & \multicolumn{1}{c|}{94.67 $\pm$ 0.07} &  \multicolumn{1}{c}{-}  & \multicolumn{1}{c}{-}  & \multicolumn{1}{c}{74.97 $\pm$ 0.16 }                       \\ 

\multicolumn{1}{l}{Random Erasing~\cite{zhong2020random}}  & \multicolumn{1}{c}{ 94.34 $\pm$ 0.10 }  & \multicolumn{1}{c}{ 94.87 $\pm$ 0.09}  & \multicolumn{1}{c|}{95.11 $\pm$ 0.07} &  \multicolumn{1}{c}{72.82 $\pm$ 0.32}  & \multicolumn{1}{c}{75.71 $\pm$ 0.16}  & \multicolumn{1}{c}{76.31 $\pm$ 0.33}                             

\\

\multicolumn{1}{l}{Hide-and-seek~\cite{kumar2017hide}}& \multicolumn{1}{c}{-}  & \multicolumn{1}{c}{94.97$\pm$ 0.00}  & \multicolumn{1}{c|}{\textcolor{blue}{\textbf{95.41 $\pm$ 0.00}}}    & \multicolumn{1}{c}{-}  & \multicolumn{1}{c}{75.82 $\pm$ 0.00 }  & \multicolumn{1}{c}{76.47 $\pm$ 0.00}                                     \\  

\multicolumn{1}{l}{GridMask~\cite{chen2020gridmask}}   & \multicolumn{1}{c}{94.60 $\pm$ 0.12}  & \multicolumn{1}{c}{94.72 $\pm$ 0.13}  & \multicolumn{1}{c|}{95.11 $\pm$ 0.21} & \multicolumn{1}{c}{72.01 $\pm$ 0.12}  & \multicolumn{1}{c}{75.13 $\pm$ 0.12}  & \multicolumn{1}{c}{75.43 $\pm$ 0.21}                       \\

\multicolumn{1}{l}{Ours} & \multicolumn{1}{c}{\textcolor{blue}{\textbf{94.65 $\pm$ 0.21}}}  & \multicolumn{1}{c}{\textcolor{blue}{\textbf{94.88 $\pm$ 0.05}}}  & \multicolumn{1}{c|}{95.14 $\pm$ 0.10}   &    \multicolumn{1}{c}{\textcolor{blue}{\textbf{73.37 $\pm$ 0.20}}}  & \multicolumn{1}{c}{\textcolor{blue}{\textbf{75.83 $\pm$ 0.21}}}  & \multicolumn{1}{c}{\textcolor{blue}{\textbf{76.84 $\pm$ 0.10}}}    \\ \bottomrule

\end{tabular}

\end{table*}

\subsection{Trade-off  between accuracy vs bias}
{ The trade-off between accuracy and bias in our proposed method reveals that while a significant reduction in bias is achieved, as illustrated in Table~\ref{tab:gender_bias}, the accuracy performance is also competitive. In Table~\ref{CIFAR_accuracies}, we observe that our method exhibits notable accuracy across various architectures in the CIFAR10 and CIFAR100 datasets, and most of the models demonstrate superior performance compared to existing methods. However, this competitive performance is not uniform, as there is one specific instance where the accuracy was lower than the baseline. This indicates that while our saliency-based masking technique effectively mitigates bias, it may introduce slight fluctuations in accuracy in certain contexts, necessitating further investigation into the balance between bias reduction and model performance. Moreover, it shows that the accuracy and bias trade-off also depends on the nature of the data set, model, and application context. }

\subsection{Real-World Application}\label{real_world_application}
{
The FaceSaliencyAug method offers practical solutions for reducing bias in various computer vision models, with key applications including:

\textbf{Facial Recognition Systems:} Common in security and surveillance, facial recognition systems often exhibit demographic biases, leading to errors like marking male instead of female or white instead of non-white. FaceSaliencyAug can help reduce these biases, ensuring more accurate and inclusive recognition across diverse groups~\cite{buolamwini2018gender, norori2021addressing}.

\textbf{Healthcare:} In facial-based diagnostics, biased models may lead to misdiagnosis. FaceSaliencyAug can help ensure healthcare models are more equitable by addressing biases related to gender and ethnicity~\cite{norori2021addressing}.

\textbf{Human Resources :} Automated HR tools, like resume screening and facial analysis in interviews, may perpetuate gender biases. FaceSaliencyAug can promote fairness in hiring by minimizing these biases.}

\section{Conclusion}

We introduced FaceSaliencyAug, a new method aimed at diversifying data to lessen geographical and stereotypical biases and diminish gender bias in CNNs and {ViTs}. By leveraging salient regions detected by our Saliency, we implemented a unique data augmentation technique that randomly masks these regions and restores the original image, thereby increasing data diversity and reducing biases. Our experiments across various datasets, including FFHQ, WIKI, IMDB, LFW, UTK Faces, and Diverse Dataset, showed improved diversity metrics, as evaluated by ISS$_{intra}$ and ISS$_{inter}$ algorithms. Additionally, our method effectively reduced gender bias in datasets such as CEO, Engineer, Nurse, and School Teacher, as indicated by reductions in IIAS for both CNNs and ViTs. {The proposed augmentation has also shown impressive performace on CIFAR10  and CIFAR100 across various architectures.}Overall, our research emphasizes the importance of addressing biases in computer vision models for fairness and inclusivity. By introducing innovative data augmentation, we've demonstrated significant enhancements in dataset diversity and reductions in gender bias across various occupations, showing promise for real-world applications of computer vision systems.

\section{Acknowledgments}

This research was supported by Science Foundation Ireland under grant numbers 18/CRT/6223 (SFI Centre for Research Training in Artificial intelligence), SFI/12/RC/2289/$P\_2$ (Insight SFI Research Centre for Data Analytics), 13/RC/2094/$P\_2$ (Lero SFI Centre for Software) and 13/RC/2106/$P\_2$ (ADAPT SFI Research Centre for AI-Driven Digital Content Technology). For the purpose of Open Access, the author has applied a CC BY public copyright licence to any Author Accepted Manuscript version arising from this submission.

\bibliographystyle{plain}
\bibliography{sn-bibliography}

\end{document}